# A Collaborative Ecosystem for Digital Coptic Studies

**Caroline T. Schroeder[1*], Amir Zeldes[2]**

1 University of Oklahoma, United States of America

2 Georgetown University, United States of America

*Corresponding author: Caroline T. Schroeder ctschroeder@ou.edu

**Abstract**
Scholarship on underresourced languages bring with them a variety of challenges which make access to the full spectrum of source materials and their evaluation difficult. For Coptic in particular, large scale analyses and any kind of quantitative work become difficult due to the fragmentation of manuscripts, the highly fusional nature of an incorporational morphology, and the complications of dealing with influences from Hellenistic era Greek, among other concerns. Many of these challenges, however, can be addressed using Digital Humanities tools and standards. In this paper, we outline some of the latest developments in Coptic Scriptorium, a DH project dedicated to bringing Coptic resources online in uniform, machine readable, and openly available formats. Collaborative web-based tools create online 'virtual departments' in which scholars dispersed sparsely across the globe can collaborate, and natural language processing tools counterbalance the scarcity of trained editors by enabling machine processing of Coptic text to produce searchable, annotated corpora.

## INTRODUCTION

Small but data-rich fields of research bring with them a variety of challenges which make access to the full spectrum of source materials and their evaluation difficult. In the case of Coptic Studies, the language and literature of this area of scholarship provide key source material for many disciplines, including linguistics, history, religious studies, and classics. Coptic is the last phase of the ancient Egyptian language family, a language that came into use in the Roman period of Egypt's history and derives from the more ancient language of the hieroglyphs. Together with Ancient Egyptian, it forms the longest continuously attested language of humanity. Yet despite its importance, few departments have scholars of Coptic, and with rare exceptions, almost never more than one.

Stable standard resources with established funding structures do not exist as they might for 'larger' ancient languages such as Latin or Greek, much less modern languages such as English. Individual scholars often produce valuable renditions of primary resources, which are published in print editions. With the exception of documentary papyri—wills, letters, testaments, receipts—digital editions of Coptic texts are rare, and historically have followed idiosyncratic digital and editorial standards. The heterogeneous nature of this digital textual data complicates natural language processing and even basic search, since Coptic grammar is fusional, meaning that basic forms or words can only be found after a standardized grammatical analysis has been carried out. Additionally, due to the colonial history of Egypt, many Coptic texts are unpublished, fragmentary, or dismembered—preserved fragment by fragment in different libraries around the globe.

Many of these challenges can be addressed using Digital Humanities tools and methods. The material and human conditions of Coptic literature and Coptic studies make a virtual research environment optimal for ongoing scholarship. Coptic Scriptorium provides such a platform. Located at copticscriptorium.org, it is an interdisciplinary, collaborative digital project dedicated to bringing Coptic cultural heritage resources online in machine readable and openly available formats [http9; Schroeder & Zeldes, et al., 2013-].[1] Our latest work encompasses four domains:

1. Collaborative annotation tools, which allow real time sharing of work in 'virtual departments' larger than any existing Coptic Studies program at a physical institution.
2. Natural Language Processing tools for Coptic, which mitigate the dearth of skilled annotators available for the vast amounts of materials that would otherwise be unsearchable, and address issues in consistency of standards and access to the otherwise opaque morphology of the language.
3. Data models for search, visualization, archival, and citation of Coptic materials which are machine actionable and make online resources usable in dependable ways.
4. Producing open, linkable data for a growing digital ecosystem in Coptic studies and the larger context of digital humanities resources for the ancient world

This paper focuses on areas 1 & 2, demonstrating how a small field for an under-resourced language can leverage diverse, interdisciplinary methods to produce open corpora for research and cultural heritage preservation.

**I THE NEED FOR A MULTIDISCIPLINARY DIGITAL COPTIC RESEARCH ENVIRONMENT**

As the language of late antique and early Byzantine Egypt—a region with a dry climate optimal for preserving textual records on many media (papyrus, parchment, stone, clay, etc.)—Coptic provides a data-rich field for researchers interested in studying multiple aspects of premodern societies. Letters, wills, and receipts testify to daily life as well as local and regional economies and governance structures. Saints' lives, texts from and about the bible, documentary sources, inscriptions, magical spells, treatises, and monastic rules illuminate the religions of Christians, Jews, Muslims, and practitioners of traditional Egyptian religion. As the last phase of the Egyptian language family, Coptic literature provides data points for studying language change that occurred over thousands of years. The alphabet used to record Egyptian moved from the ancient hieroglyphs used for thousands of years in ancient Egypt to a script known as Demotic beginning in 650-400 BCE to Coptic in the Roman era; in contrast to the glyphs of ancient Egyptian, the Coptic alphabet uses the ancient Greek alphabet plus a handful of additional characters from Demotic. As a result, Coptic gives us the most detailed window of any era into how the Egyptian language was pronounced, details which are much more obscure and contested for earlier periods due to the partial representation of phonological forms in the hieroglyphic script. Coptic grammar evolved from Demotic and ancient Egyptian; much of the vocabulary is Egyptian with additional words from contact languages (especially, but not only, Greek) incorporated over time. Thus, the same texts provide important primary sources for multiple academic disciplines.

---

[1] The Coptic Scriptorium team thanks the following institutions for supporting our work: the National Endowment for the Humanities (PW-51672-14, HD-51907-14, HG-229371, HAA-261271-18), Deutsche Forschungsgemeinschaft, Humboldt University, Georgetown University, the University of the Pacific, the University of Oklahoma, the Berlin-Brandenburg Academy of Sciences and Humanities, Canisius College.

Despite its importance, Coptic shares several qualities with other under-resourced languages and has not been considered one of the primary languages for the study of the ancient Mediterranean, which typically consist of Greek, Latin, and Hebrew. (On the rich resources for digital and computational study in Greek and Latin, by comparison, see [Berti, 2019; Reggiani, 2017; Reggiani, 2018].) As with other under-resourced languages, the challenges facing research in Coptic have multiple origins—some linguistic, some historical, some political. Its syntax and morphology differ from other major language systems. One cannot so easily apply principles from a more common related language to learning or researching Coptic, as one can within language groups like Semitic languages or Romance languages. Few native speakers and readers of Coptic remain among Coptic Orthodox Christians in Egypt or the Coptic Diaspora, which means that cultural heritage literature written in the Coptic language is no longer accessible to most members of the heritage group. As a common, spoken language in Egypt, Coptic was overtaken by Arabic. The church continues to use Bohairic, one of the main dialects of Coptic, in liturgy, and heritage organizations (such as the Saint Shenouda the Archimandrite Coptic Society in Los Angeles) promote the study of Coptic language, literature, and history among Copts. While the Saint Shenouda Society participated in early Coptic digitization efforts during the 20th century and distributed the Coptic New Testament and a few other works to Society members and interested scholars who purchased their CD-ROMs, the scope was limited. [Schroeder 2019; http6] Presently, Coptic typically is no longer spoken in homes, schools, and other spaces that encourage oral and written language preservation across generations. Thus, literature written in the Coptic language has been readable only to few people, and until recently without access to basic digital tools, such as morphological analysis and a linked online dictionary, both of which Coptic Scriptorium helps to develop.

Additionally, many of the cultural heritage documents in question are no longer in possession of the heritage group, or even accessible to most Copts. Due to the history of colonialism and the antiquities trade in Egypt in the 17th century onward, Coptic literary documents have been dispersed across the globe, often with pages from the same codex (or ancient book) in multiple libraries or collections. [Schroeder & Zeldes, 2016b] Moreover, not all of these collections are housed in academic or public libraries and museums; some Coptic manuscripts are in the hands of private collectors. As an example, we can examine one sermon or treatise known as 'I See Your Eagerness.' A famous monk and monastic leader named Shenoute wrote the work sometime in the 400s CE, and the text was copied by subsequent generations of monks at his monastery, now known as the White Monastery near Sohag, Egypt. Today, four copies of the ancient book containing 'I See Your Eagerness' survive, all in fragments. [Emmel, 2004:1:255-69, 2:628-32, 825–27] These four surviving codices likely date to the tenth or eleventh centuries.  Additionally, one surviving copy of a lectionary codex also contains an excerpt of the text. [Emmel, 2004: 1:362-368] The five known, extant codices exist in fragments with pages distributed across nine institutions: the British Library (London), the Österreichische Nationalbibliothek (Vienna), the Coptic Museum (Cairo), the Bibliothèque nationale (Paris), the John Rylands University Library (Manchester), the Biblioteca Nazionale 'Vittorio Emanuele III' (Naples), the Rijksmuseum van Oudheden (Leiden), the Bodleian Library (Oxford), and the Reale Accademia Nazionale dei Lincei (Rome). [Emmel, 2004: 1:255-69, 397, 399, 426–28, 435–36, 469-70] 'I See Your Eagerness' is just one literary work witnessed in these codices; the pages containing its text are distributed in Paris, Naples, Vienna, and Manchester. Patching together the folios of each ancient codex still leaves gaps, holes in the text that no known copy fills. Furthermore, not all of the known, extant pages have been published in print, and only in 2015 was an English translation of the text finally published. [Brakke & Crislip, 2015:91-105] Only on the Coptic Scriptorium website can one find and read the majority of the Coptic text of this work in one place in sequence, complete with grammatical analyses which allow readers to find specific words linked to dictionary entries, abstracted from the inflected and fused forms which they take on in the text. [Shenoute]

Prior to the launch of Coptic Scriptorium, three major digital resources for literary Coptic texts existed, each making important advances in the field, although none providing a collaborative environment for digitization, annotation, and open access publication. The Packard Humanities Institute (PHI) released a CD of the Coptic New Testament and texts from the Nag Hammadi library (a collection of fourth century Coptic codices discovered in 1945). PHI texts circulated, were re-edited, and redistributed in various formats. The Corpus dei Manoscritti Copti Letterari (CMCL) project hosts photographs of manuscripts, transcriptions of texts, and identifiers for codices, works, and authors (among many other resources). [http1] The Marcion Project digitized the text of several out of copyright print editions as well as existing digital resources (such as texts originating from PHI), linked them to lexical resources, and created a software package that researchers can download to read and use the resources. [http2] (For a more detailed history of early Coptic Digital Humanities see [Schroeder, 2019].) In addition, Papyri.info has published a significant number of digital editions of Coptic documentary papyri and ostraca, which contain important records of daily life, though not literature. [http3; Sosin, 2010; Vannini, 2018]

By comparison, the ancient language Latin has substantial digital resources, including, for example, the new *Digital Latin Library* sponsored in part by the American academic professional society for Classics (the Society of Classical Studies). The *Digital Latin Library* provides a digital ecosystem for cataloguing Latin texts and publishing digital editions with robust metadata, data visualizations, and an interactive critical apparatus to compare variants of a text. These features provide new infrastructure for Latin that builds previous work by DH projects in Latin stretching back decades. [Huskey, 2019; http13]

Coptic Scriptorium thus fills the need for a DH project that allows researchers to build collaborations larger and more diverse than any realistic, onsite collection of Coptologists at any one university, with a commitment to not just archiving facsimiles or unanalyzed keyed in texts, but also making texts searchable and accessible through linguistic analysis. As a platform, the project forms a large, virtual, interdisciplinary department with shared tools (particularly building on benefits from a growing array of natural language processing tools) and joint digital publications.

**II COLLABORATIVE ANNOTATION TOOLS**

The collaborative annotation tools we have developed leverage the knowledge base of multiple academic fields. Researchers in different disciplines will come to a document with different research questions and different methodologies. Philologists and literary specialists may be interested in reading entire texts from beginning to end and in creating digital editions of primary sources. Historians may be interested in close readings of primary documents, searching large aggregate text collections for relevant sources, and building such large digital collections to facilitate search. Linguists may be interested in researching morphology, syntax, and language change and in creating normalized, richly annotated digital text corpora to facilitate that research computationally. Scholars of religion may be interested in investigating what people wrote about certain topics (God, sexuality, ritual), analyzing how ancient writers interpreted prior religious literature, and creating digital collections that can help track connections between concepts and between texts. Consequently, our project has built tools for transcribing Coptic (including encoding paleographic and manuscript information), for segmenting, normalizing and tagging Coptic words, for annotating for linguistic information, and providing rich metadata to catalogue and contextualize documents and their history. Our collaborative annotation tools are thus multidisciplinary, enabling researchers in different fields to digitize texts important to them, to annotate the texts in ways that are meaningful for their future scholarship, and ultimately to query the corpora for their research. Importantly, we emphasize generic workflows and formats to represent annotations which allow us to generalize across those different needs without having to construct dedicated tools for each discipline or methodology.

Where possible, we adapt existing 'off-the shelf' tools built for other digital scholarly projects and have experienced success training graduate students and recent PhDs with expertise in Coptic but little prior DH experience. For example, our team members use Arborator [Gerdes, 2013; http7], an existing syntax annotation web interface, to markup Coptic text for dependency syntax annotations. Often referred to as 'treebanks', resources with such annotations document relational grammatical information such as subjects and objects of verbs, prepositional modifiers, and other syntactic relationships. [http10] These annotations enable complex linguistic research for linguists, who are often interested in the syntactic behavior of words of different classes (e.g. when can Greek nouns and verbs combine with Coptic words to form compounds?) or the circumstances distinguishing the usage of competing constructions (what is the difference between two ways of saying something?). Historians and scholars of religion can also use them to answer research questions such as how does a particular author write about or conceptualize a topic (e.g., God, children, books, the law) by enabling queries for terms dependent on or otherwise linguistically connected to their topic of interest (for example 'what verbs are demons agents of?', or 'which predicates distinguish first person objects in two genres?'). Similarly, we utilize the existing tool WebAnno [Yimam et al., 2013; http5] to annotate Coptic literature for entity types and coreference information. These annotations mark relationships between entities (persons, places, objects, etc.) and link multiple references within a text belonging to the same entity (including pronouns, alternate titles, etc.). All of these annotations are offered in a number of automatically generated formats outputted and merged by our tools via GitHub, which also maintains a complete revision history for annotations of each document that we release [http16].

In those cases where we have found it necessary to build custom tools, the new software is open source and built in conversation with other developments in DH infrastructure; we strive to keep our tools flexible so more researchers can in turn adapt them for their use. A central practice in our work is transcribing digital Coptic text from manuscript facsimiles, creating digital editions directly from the primary sources. We share this methodology of 'digital philology' with many other text-based digital humanities projects, especially those focused on the ancient and medieval literary traditions. Existing transcription tools, however, did not always meet our needs. If built for the Latin alphabet, they did not always function well with the Coptic Unicode character set. Additionally, after years of manually transcribing in text files and then manually running individual natural language processing (NLP) tools, we wanted an inclusive work environment to connect our transcription mechanism to NLP tools, one which carried over our metadata from one format (encoded transcriptions) to another (a multilayer annotation file) after linguistic annotations. (On multilayer annotation, see [Zeldes, 2018: Part I].) An infrastructure that enabled us to save new editions of the data in a version-controlled environment was also essential. Other transcription web applications which we considered using, such as T-Pen and the Papyri.info SOSOL editor, functioned well for those projects. [http4; http16] Nonetheless, we concluded that building our own tool—one the project team could manage and update easily itself and customize for our particular research questions—proved the best solution. Even in creating something anew, however, we built on top of existing open source resources: the transcription mode uses the open-source online text and XML editor CodeMirror [http8], and the open-source spreadsheet EtherCalc [http14] provides the core of the multilayer annotation editor. Using EtherCalc as an infrastructure in particular, which allows for live concurrent editing (similarly to a Google spreadsheet) means that project contributors from different departments can view and discuss each other's annotations in real time, forming a virtual working group that can grow much larger than the one or two researchers interested in Coptic which might realistically be found at most institutions.

Each of the different working modes used in our annotation environment has configurable validation options, so that a research project can ensure its data—whether XML encoded text or other types of multilayer annotations—conforms to project specifications and standards. Additionally, users can both save the data on the project's server and commit it to a version-controlled GitHub repository. Thus, our collaborative transcription and annotation tool, GitDox, was born. [Zhang & Zeldes, 2017; http11] While Coptic Scriptorium uses it for Coptic, it is adaptable to other languages and has been used for projects in English, and for non-historical data as well. Similarly, while the existing tool WebAnno works well for manual entity and coreference annotation by language experts, we needed an application for *automated* entity annotation in order to scale up annotations for people, places, and things in a large corpus. Project members therefore developed a tool called xrenner, which ingests plain text or syntactically parsed data and produces node annotations for both named and non-named entities, as well as edge annotations describing relationships between references to the same entity. [Zeldes & Zhang, 2016; http12] The tool is also configurable for multiple languages and currently has models online for Coptic, English, Hebrew, and German.

**III NATURAL LANGUAGE PROCESSING**

For complex computational research as well as for simple word searches, digital scholarship in Coptic Studies requires a robust set of natural language processing tools (NLP). Coptic is an agglutinative language, where one group of letters can consist of a chain of multiple linguistic units. For example, one might see multiple prefixes conveying tense, negation or future action, other aspects of grammar (e.g., the formation of a relative clause), and a subject pronoun followed by a verb and object affixes. Moreover, no one standard for printing or transcribing Coptic exists. [Schroeder & Zeldes, 2016a] Word searches alone require tools to segment and lemmatize Coptic text. Even though our project may maintain internal standards on transcription to ensure the tools train and run on standard, expected textual data, research partners may share digitized texts which were encoded at different times, by different teams, and using different standards. Thus, Coptic textual data often requires preprocessing to ensure higher accuracy rates for subsequent NLP tools, as described in detail in [Zeldes, 2019a; Zeldes, 2019b; Zeldes, 2019c].

Our current suite of NLP tools applies some preprocessing standardization followed by word segmentation, normalization, lemmatization, part of speech tagging, language of origin tagging, multiword expression recognition, and dependency parsing (treebanks). Researchers can download and install the toolset, enter text into the tool pipeline online on our website, or utilize our machine accessible API. [Zeldes & Schroeder, 2016; http15] Automated entity annotations, discussed above, are still under development and will be incorporated into the pipeline in future years.

These technologies produce a rich set of annotations that enable diverse and multidisciplinary forms of research. Language learners interested in reading a text can browse a visualization of the normalized text that is linked to an online dictionary. The dictionary is a product of a collaboration with multiple projects in Germany. Thus, lemmatization and part of speech tagging (methodologies central to linguistics) also facilitate close reading and language pedagogy, as well as linking to online dictionary entries. Linguists can use the part of speech, treebank, and language of origin annotations to research syntax, morphology, and Egyptian language contact with Greek. Historians can conduct topical research using basic queries or more complex historical text analysis using the normalization, lemma, and treebanking annotations.

All of this work occurs in conversation with wider Digital Humanities and Computational Linguistics research. We are building our Coptic treebank within the framework of the Universal Dependencies project (UD). [Zeldes & Abrams, 2018; http17] The UD project aggregates over 100 treebank corpora from over 70 languages according to a common standard, in order to facilitate cross-language linguistic research. We draw on the standardization work they have already conducted so that Coptic Studies, a small field, doesn't devote resources to 'reinventing the wheel,' (Coptic is one of a few ancient languages with treebank corpora. Several treebanks exist for the more well-resourced languages of Greek and Latin, some of which have developed according to different annotation standards and have been or are being ported to the Universal Dependencies framework, as described in [Celano, 2019].) Additionally, due to Coptic's inclusion in UD, parsers and technologies developed for other languages will be evaluated on Coptic. This brings an under-resourced language into dialogue with other language systems and promotes the availability of more tools for the language, not necessarily because they are targeting Coptic, but because they support 'all UD lanaguages'. [Pinter et al., 2019], for example, recently conducted a cross-language study of part-of-speech tagging methods that included Coptic.

**CONCLUSIONS**

An interdisciplinary, digital ecosystem thus aggregates scholarly resources—human, linguistic, and technological— in one place. Pairing Digital Humanities tools and methods with natural language processing results in a rich environment for diverse research in Coptic Studies. Much of the work we have described involves adapting or building tools. In recent years, digital humanists such as [Burdick et al., 2012:122; Ramsay & Rockwell, 2012; Ramsay, 2016; Golumbia, 2019] have engaged in self-reflection on the comparative value of tool-building, project creation, and other modes of Humanities scholarship: is tool-building itself the production of knowledge? Do tools express arguments, or are they methods and vehicles for the subsequent production of knowledge? These disciplinary boundary questions fall to the shadows when we center the production of knowledge on under-resourced languages and cultures. We build when we need something different from existing infrastructures, and we adapt and reuse whenever possible, since we do not have extensive human or financial resources. Our tools indeed produce arguments, arguments about the vitality of interdisciplinary collaboration and questions of detail that would not be raised and treated explicitly if we were not forced to make our assumptions, analyses, and guidelines explicit in the form of digital resources.

Miram Posner in [Posner, 2019] has argued for the importance of building community in DH—particularly developing local communities—and for being attendant to the needs of those local communities. She urges us to prioritize 'a community of people who learn together, support each other, and trust each other' over the traditional DH outputs of projects and tools. Researchers and heritage readers of under resourced languages particularly feel the urgency of community building. While Posner refers to local communities on college and university campuses, we take as our 'local' ecosystem our small Coptic Studies community worldwide, which is often dispersed, with one person at any given campus. We build tools and develop projects, but in the service of creating and supporting our virtual local community which in turn makes the project grow, evolve, and make more resources for Coptic available for everyone.